%% file: main.tex
\documentclass{article}

\usepackage[main, final, nonatbib]{neurips_2026}

\usepackage[utf8]{inputenc} % allow utf-8 input
\usepackage[T1]{fontenc}    % use 8-bit T1 fonts
\usepackage{hyperref}       % hyperlinks
\usepackage{url}            % simple URL typesetting
\usepackage{booktabs}       % professional-quality tables
\usepackage{amsfonts}       % blackboard math symbols
\usepackage{nicefrac}       % compact symbols for 1/2, etc.
\usepackage{microtype}      % microtypography
\usepackage{xcolor}         % colors
\usepackage{amsmath}

\usepackage{multirow}
\usepackage{graphicx}

\usepackage{amssymb}
\usepackage{textcomp}
\usepackage{gensymb}
\usepackage{wrapfig}
\usepackage{subcaption}
\usepackage{pifont}

\usepackage[ruled,vlined,linesnumbered]{algorithm2e}
\usepackage{tabularx}

\SetKwComment{tcp}{\textcolor{gray}{$\triangleright$~}}{}

\newcommand{\I}{\mathsf{I}}

\title{Guiding Visual Autoregressive Models through Spectrum Weakening}

\author{%
  Chaoyang Wang \qquad Tianmeng Yang \qquad Yunhai Tong \\
  State Key Lab of General AI, School of Intelligence Science and Technology, Peking University \\
  \texttt{cywang@stu.pku.edu.cn} \\
}

\begin{document}

\maketitle

\input{sec/0_abstract}    
\input{sec/1_intro}

\input{sec/2_related_work}

\input{sec/3_method}
\input{sec/4_exp}

\input{sec/5_conclusion}

{\small
\bibliographystyle{ieee_fullname}
\bibliography{refbib}
}

\input{sec/6_suppl}

\end{document}

%% file: sec/0_abstract.tex
\begin{abstract}

Classifier-free guidance (CFG) has become a widely adopted practice for enhancing generation quality and improving condition alignment. Recent studies have explored guidance mechanisms for unconditional generation, yet these approaches remain fundamentally tied to assumptions specific to diffusion models.
In this work, we propose a spectrum weakening framework for visual autoregressive (AR) models. The method is training-free and condition-free, constructing a controllable weak model in the spectral domain without architectural modifications. Specifically, we selectively retain spectral components along the channel dimension of internal representations to weaken features without requiring a complete spatial layout. We further apply spectrum renormalization to maintain feature scales during generation.
Comprehensive empirical and ablation studies confirm the effectiveness of our approach, including raster-scan, random-order, and scale-wise AR, spanning discrete and continuous modeling and covering class and text-condition scenarios, demonstrating high-quality unconditional generation and strong prompt alignment for conditional generation.
Project Page: \url{https://wang-chaoyang.github.io/project/swg}.
\end{abstract}

%% file: sec/1_intro.tex
\section{Introduction}
\label{sec:intro}

Autoregressive (AR) modeling has recently emerged as a promising paradigm in visual generation, demonstrating remarkable scalability and controllability across a wide range of downstream applications, including video synthesis~\cite{agarwal2025cosmos,deng2024autoregressive}, artistic creation~\cite{sun2024autoregressive}, and high-fidelity image generation~\cite{li2024autoregressive}. 
Unlike diffusion models~\cite{song2020denoising,song2020score,ho2020denoising} that iteratively refine noisy images as a whole, visual AR models construct images token by token, predicting each part conditioned on previously generated content.

Despite their success, existing visual generation models~\cite{esser2021taming,esser2024scaling,karras2024analyzing,podell2023sdxl,Peebles_2023_ICCV} often depend on classifier-free guidance (CFG)~\cite{ho2022classifier}, which requires learning unconditional predictions alongside conditional generation.
However, this requirement constrains the training setup, limits applicability to models without unconditional training, and does not directly provide guidance for unconditional generation.
Recently, inspired by contrastive decoding~\cite{li2023contrastive}, several works have proposed guiding the generation process using a weaker model~\cite{karras2024guiding} rather than an explicit classifier. 
These methods have shown strong performance in diffusion models, typically by modifying network architectures~\cite{hyung2025spatiotemporal}, designing perturbed timesteps~\cite{sadat2024no}, or perturbing self-attention maps~\cite{hong2024smoothed, hong2023improving, ahn2024self}. 
However, as shown in Tab.~\ref{tab:paradigm_compare}, these strategies are challenging to transfer to general AR generation, which has no notion of diffusion timesteps and does not form a coherent global image representation until the entire sequence is decoded.

\begin{figure}[t!]
    \centering
    \includegraphics[width=1.0\textwidth]{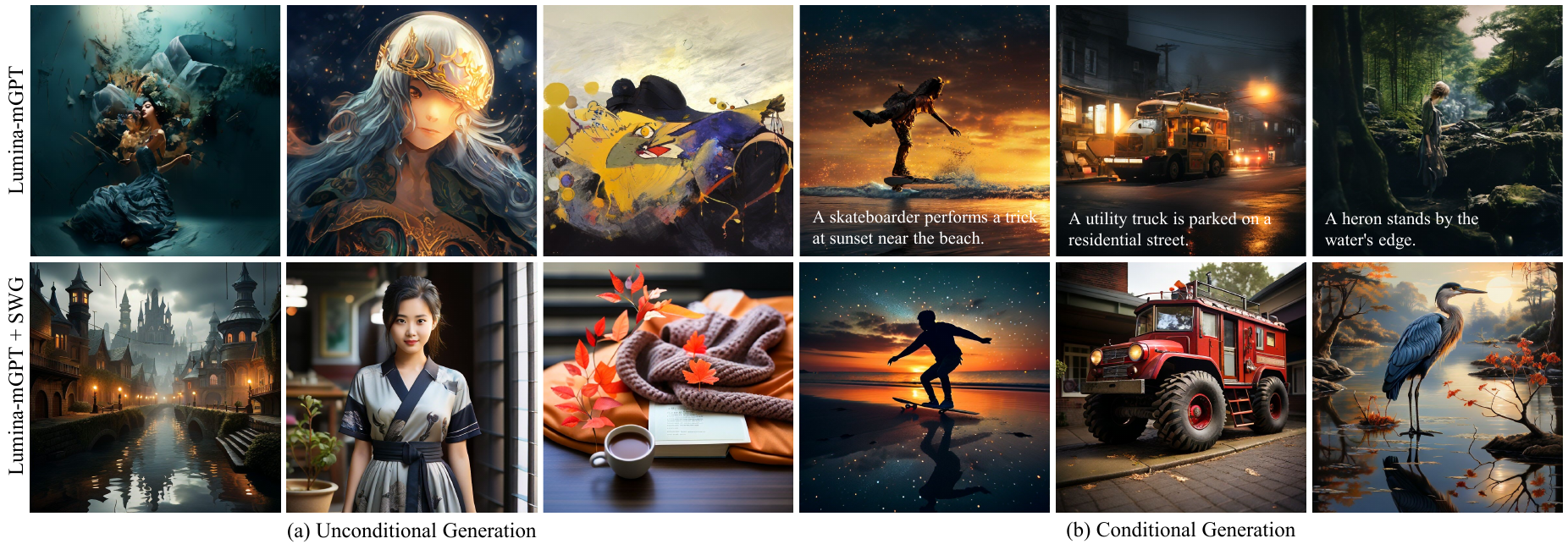}
    \caption{Visual results of Spectrum Weakening Guidance (SWG) on Lumina-mGPT. (a) The model synthesizes images using a null prompt. (b) The model synthesizes images using the specified prompts, shown as white text in the first row.}
    \label{fig:main_lumina}
\end{figure}

In light of these challenges, we propose a training-free spectrum weakening guidance (SWG) approach for visual AR generation. 
Specifically, our method has four key properties: 
1) no reliance on external conditions, naturally supporting unconditional generation; 
2) preservation of the original architecture, with no structural changes or retraining; 
3) no assumption of a complete global layout, fitting AR decoding with only partial context; 
and 4) enabling flexible adjustment of guidance intensity. 
As shown in Fig.~\ref{fig:main_lumina}, our proposed SWG significantly improves visual quality in both conditional and unconditional generation.

Concretely, we apply the discrete Fourier transform (DFT) along the channel dimension of intermediate features and use a binary mask to selectively retain spectral components. This provides a flexible way to construct weakened representations while preserving their shape and token layout. Since spectrum selection can alter feature magnitudes, we apply a renormalization step to maintain feature scales during generation.
Empirically, we broadly compare and validate the effectiveness of our SWG, including raster-scan, random-order, and scale-wise AR, spanning discrete and continuous modeling and covering class and text-condition scenarios, and observe consistent quality gains. 
Further ablation studies demonstrate that the constructed weak model produces more uncertain predictions, and outperforms alternative weak-branch constructions.

The main contributions are as follows: 1) In this paper, we investigate and propose a training-free and condition-free weak model guidance approach for visual AR generation. 2) We design a spectrum selection pipeline to construct weakened intermediate representations, with renormalization to maintain feature scales during generation. 3) Comprehensive comparative and ablation experiments demonstrate the encouraging effectiveness and generalization of our method, as well as flexible controllability and compatibility with existing guiding techniques.

%% file: sec/2_related_work.tex
\section{Related Work}
\label{sec:related_work}

\noindent
\textbf{Visual Autoregressive Model} is motivated by the success of large language models (LLMs)~\cite{brown2020language,touvron2023llama} and has recently emerged as a promising alternative to diffusion models because of its strong scalability~\cite{tian2024visual}. 
Unlike diffusion models~\cite{lipman2022flow,liu2022flow,liu2023instaflow,podell2023sdxl,rombach2022high,song2020denoising,song2020score} that iteratively refine a complete image from noise, AR models generate visual tokens sequentially, conditioning each prediction on previously generated tokens. 
Existing approaches can be broadly divided into raster-scan models~\cite{liu2024lumina,sun2024autoregressive,ramesh2021zero,ding2021cogview,ding2021cogview2,yu2022scaling,wang2024loong,yan2021videogpt,kondratyuk2023videopoet,nash2022transframer}, which decode images in spatial order, and set- or scale-wise models~\cite{li2024autoregressive,deng2024autoregressive,chang2023muse,hong2022cogvideo,yu2023magvit,villegas2022phenaki}, such as masked or hierarchical AR variants. 
Despite their success, current AR frameworks still rely heavily on CFG and remain limited in unconditional generation. 
In this work, we revisit this limitation through the lens of weak model guidance and introduce a training-free spectrum-weakening strategy tailored for AR generation.

\begin{table}[t]
\small
\setlength{\tabcolsep}{2.8pt}

  \centering
  \caption{Comparison of training-free self-guidance methods across generation paradigms. Existing approaches are designed for diffusion models or limited to scale-wise AR. Our proposed SWG supports all three AR paradigms.}
  \label{tab:paradigm_compare}
  \begin{tabular}{l cccccc cc}
    \toprule
    & \multicolumn{6}{c}{\textit{Diffusion models}}
    & \multicolumn{2}{c}{\textit{Visual autoregressive}} \\
    \cmidrule(lr){2-7} \cmidrule(lr){8-9}
    & SAG~\cite{hong2023improving} & PAG~\cite{ahn2024self} & SEG~\cite{hong2024smoothed} & TSG~\cite{sadat2024no} & STG~\cite{hyung2025spatiotemporal} & TPG~\cite{rajabi2025token} & SSG~\cite{shin2026ssg} & SWG (Ours) \\
    \midrule
    Raster-scan AR & \texttimes & \texttimes & \texttimes & \texttimes & \texttimes & \texttimes & \texttimes & \checkmark \\
    Random-order AR  & \texttimes & \texttimes & \texttimes & \texttimes & \texttimes & \texttimes & \texttimes & \checkmark \\
    Scale-wise AR & \texttimes & \texttimes & \texttimes & \texttimes & \texttimes & \texttimes & \checkmark & \checkmark \\
    \bottomrule
  \end{tabular}
\end{table}

\noindent
\textbf{Weak Model Guidance} has recently attracted growing attention as an alternative to CFG, offering a more flexible and interpretable way to control generative models.
Existing approaches to weakening model alignment can be broadly grouped by their underlying mechanisms.
Some methods derive guidance by leveraging alternative training schedules or conditioning mechanisms. For example, Autoguide~\cite{karras2024guiding} uses states at varying convergence levels, while CFG~\cite{ho2022classifier} relies on jointly learned unconditional predictions. However, these methods require modified training procedures and precise computational control, and CFG is inherently limited to conditional setups, as the guidance process relies entirely on a conditioning signal.
Others~\cite{hyung2025spatiotemporal} adjust the model structure by layer pruning. 
In addition, several methods exploit characteristics unique to diffusion models, such as perturbing self-attention maps spatially~\cite{ahn2024self,hong2023improving,hong2024smoothed}, shuffling image tokens~\cite{rajabi2025token}, or perturbing timesteps~\cite{sadat2024no}. However, these designs rely on the full-image generation and timestep formulations specific to diffusion models, making them unsuitable for AR generation.

In this work, we revisit alignment weakening from an information-reduction perspective, applying spectrum selection to attenuate model alignment without modifying the architecture.

\noindent
\textbf{Spectral Representations} offer a powerful perspective for processing visual signals in the frequency domain. 
The transformation between spatial and spectral domains, typically implemented via the DFT, enables bidirectional conversion without significant information degradation. 
In the spectral domain, low-frequency components capture coarse structures and overall geometry, while high-frequency components encode fine details and textures, providing an interpretable decomposition of visual information~\cite{wang2025explore}. 
This property has been exploited in various diffusion-based generation and control frameworks~\cite{si2024freeu} to enhance structure–detail balance. 
Beyond interpretability, spectral representations also enable the decoupling of distinct signal components, inspiring explorations in AR modeling~\cite{yu2025frequency}. 
In contrast, we operate in the spectral domain for flexible information reduction. Spectral selection suppresses selected components while preserving feature shapes and token positions, providing a controllable alternative to spatial perturbations.

%% file: sec/3_method.tex
\section{Method}
\label{sec:method}

\subsection{Preliminary}
\label{sub_sec:preliminary}

\noindent
\textbf{Visual Autoregressive Model.}
In an AR framework, the conditional distribution of an image $x$ given a condition $c$ is factorized as:
\begin{equation}
\label{eq:ar}
p(x|c) = \prod_{t=1}^T p(x_t | x_{<t}, c),
\end{equation}
where $x_t$ denotes the $t$-th token in the image sequence.
The model is optimized by maximizing the likelihood.

To incorporate conditioning signals, such as text prompts, the model encodes the corresponding tokens and concatenates them with image tokens, enabling it to attend to both modalities jointly. 
During inference, the KV-cache stores the key-value representations of previously generated tokens, enabling efficient computation as the query $Q$ attends to the current step while $K$ and $V$ aggregate information from both past and current tokens.

CFG is also commonly used in visual AR models. An unconditional branch is trained with null prompts, and during generation, the logits from the conditional and unconditional branches are linearly combined to balance fidelity and diversity.

\noindent 
\textbf{Spatial to Spectral Transformation.} 
Let $\mathbf{x} \in \mathbb{C}^N$ denote a discrete spatial-domain signal and let
$\mathbf{\hat{x}} \in \mathbb{C}^N$ denote its discrete Fourier transform spectrum.
The discrete Fourier transform can be written in matrix form as
\begin{equation}
\label{eq:dft}
\mathbf{\hat{x}} \;=\; \mathbf{W}\,\mathbf{x},
\end{equation}
where the DFT matrix $\mathbf{W}\in\mathbb{C}^{N\times N}$ has entries
\begin{equation}
\label{eq:dft_matrix_single}
W_{k,n} \;=\; \omega^{k n}/\sqrt{N}, \qquad k,n = 0,1,\dots,N-1,
\end{equation}
and
\begin{equation}
\omega \;\triangleq\; e^{-\,\mathrm{i}\,2\pi / N}
\end{equation}
is the primitive $N$-th root of unity (here $\mathrm{i}=\sqrt{-1}$).

The inverse discrete Fourier transform (IDFT) under the common normalization convention is
\begin{equation}
\label{eq:idft}
\mathbf{x} \;=\;\mathbf{W}^{*}\,\mathbf{\hat{x}},
\end{equation}
where $\mathbf{W}^{*}$ denotes the conjugate transpose of $\mathbf{W}$. 

Specifically, the DFT matrix $\mathbf{W}$ satisfies $\mathbf{W}^{*}\,\mathbf{W} = \,\mathbf{I}$, where $\mathbf{I}$ is the $N\times N$ identity matrix.

\subsection{Spectral Weakening Guidance}

To derive a weaker variant of the original model, we attenuate the information content of its internal representations via spectral selection.
We first outline the overall pipeline and then discuss the effect of spectral selection from an information-theoretic perspective.
Finally, we introduce a spectrum renormalization step to maintain feature scales and improve numerical stability during generation.

\subsubsection{Selective Spectrum Preservation} 
We introduce a spectrum selection pipeline to modulate the information within visual AR models. 
Given feature tensor $\mathbf{x} \in \mathbb{R}^{C}$, 
we apply DFT along the channel dimension to obtain its spectral representation $\hat{\mathbf{x}} = \mathbf{W}\,\mathbf{x}$ following Eq.~\ref{eq:dft}. 
We employ a binary spectral mask to select a subset of spectral components, which is equivalently represented as a diagonal selection matrix 
\begin{equation}
\label{eq:sel_matrix}
\mathbf{M} = \mathrm{diag}(M_0, M_1, \dots, M_{C-1}) \in \{0,1\}^{C\times C},
\end{equation}
where $M_i = 1$ indicates that the $i$-th frequency is preserved, and $M_i = 0$ indicates suppression.

The filtered spectrum and its inverse transform are given by
\begin{equation}
\label{eq:apply_dft}
    \hat{\mathbf{x}}' = \mathbf{M}\hat{\mathbf{x}}, 
    \qquad 
    \mathbf{x}' = \mathbf{W}^{*}\hat{\mathbf{x}}'.
\end{equation}

Since $\mathbf{x}$ is real-valued, its Fourier spectrum $\hat{\mathbf{x}}$ exhibits conjugate symmetry, 
and numerical computation may introduce imaginary residuals. 
Therefore, we take the real part of the reconstructed signal as the final output:
\begin{equation}
\label{eq:get_real}
    \mathbf{x}' \leftarrow \Re(\mathbf{x}').
\end{equation}

The resulting $\mathbf{x}'$ retains the same shape as $\mathbf{x}$ while suppressing selected spectral components. Spectral selection acts along the channel dimension through a fixed binary mask, yielding a linear and differentiable transformation of the input and allowing controllable feature perturbations while preserving token positions.

\noindent
\textit{Discussion.} The proposed spectrum selection pipeline is flexible, as it does not rely on the availability of the complete image structure during generation. 
Furthermore, contemporary AR models typically rely on the KV cache for inference, in which each block has access to only one visible token. Consequently, structural perturbations such as patch shuffling~\cite{rajabi2025token} are infeasible in this setting.

\subsubsection{Information Reduction Analysis}

Spectral selection reduces the information available
to subsequent layers by suppressing selected frequency
components. Since the DFT is invertible, transforming
features into the spectral domain does not alter their
information content. In contrast, applying a binary
spectral mask discards selected coefficients.

Formally, let $\mathbf{Z}$ be any random variable
potentially dependent on $\mathbf{x}$. Since the
spectrally filtered representation $\mathbf{x}'$
is a deterministic function of $\mathbf{x}$,
the data processing inequality gives
\begin{equation}
\label{eq:information_reduction}
\I(\mathbf{x}';\mathbf{Z})
\leq
\I(\mathbf{x};\mathbf{Z}).
\end{equation}

This indicates that spectral selection cannot increase the information about $\mathbf{Z}$ contained in the original representation. By controlling the preserved spectral components, we obtain a simple mechanism for constructing weakened representations for guidance.

\begin{algorithm}[t]
\setlength{\algomargin}{3pt}%
\SetAlgoNoEnd            % 
\SetAlgoNlRelativeSize{-1}%
\SetAlgoInsideSkip{medskip}%
\DontPrintSemicolon
\caption{Sampling with Spectrum Weak Guidance}\label{alg:swg}
\KwIn{SWG scale $\omega_s$; CFG scale $\omega_c$ (opt.); binary spectrum mask $\mathbf{M}$; condition $c$; SWG module set $\mathcal{S}\!\subseteq\!\theta\!=\!\{\theta_1,\dots,\theta_N\}$}
\KwOut{Generated sample $x_T$}
$x_0 \leftarrow \langle\mathrm{BOS}\rangle$\;
\For{$t = 1, \ldots, T$}{
  $z_{t,c} \leftarrow f_\theta(x_{t-1}, c)$
  \tcp*[r]{Cond.\ pred.}
  $\ell_0 \leftarrow x_{t-1}$\;
  \For{$i = 1,\dots,N$}{
    $\ell_i \leftarrow f_{\theta_i}(\ell_{i-1},c)$\;
    \If{$\theta_i \in \mathcal{S}$}{
      $\hat{\ell}_i \leftarrow \mathbf{M}\mathbf{W}\ell_i$\;
        $\ell_i \leftarrow
        \mathrm{Renorm}(\hat{\ell}_i, \ell_i)$\;
    }
  }
  $z_{t,p} \leftarrow \ell_N$;~~$z_t \leftarrow z_{t,c} + \omega_s(z_{t,c} - z_{t,p})$
  \tcp*[r]{Apply SWG}
  \If(\tcp*[f]{Optional CFG}){use CFG}{
    $z_{t,b} \leftarrow f_\theta(x_{t-1}, \emptyset)$;~~$z_t \leftarrow z_t + \omega_c(z_{t,c} - z_{t,b})$
  }
  $x_t \leftarrow \mathrm{Sample}(\mathrm{Softmax}(z_t))$\;
}
\Return{$x_T$}
\end{algorithm}

\subsubsection{Spectrum Renormalization}
\label{subsec:method_renorm}

Spectral selection may alter the magnitude of intermediate representations. We therefore apply a renormalization step to maintain feature scales during generation.

Specifically, given the original spectrum
$\hat{\mathbf{x}}=\mathbf{W}\mathbf{x}$
and the filtered spectrum
$\hat{\mathbf{x}}'=\mathbf{M}\hat{\mathbf{x}}$,
the renormalized representation can be obtained
in the spectral domain as
\begin{equation}
\label{eq:spec_renorm}
\tilde{\hat{\mathbf{x}}}' =
\hat{\mathbf{x}}' \cdot
\frac{\|\hat{\mathbf{x}}\|_2}
{\|\hat{\mathbf{x}}'\|_2 + \epsilon},
\quad
\tilde{\mathbf{x}}' =
\Re\big(\mathbf{W}^{*}\tilde{\hat{\mathbf{x}}}'\big),
\end{equation}
or alternatively, after reconstruction in the
spatial domain as
\begin{equation}
\label{eq:spat_renorm}
\tilde{\mathbf{x}}' =
\mathbf{x}' \cdot
\frac{\|\mathbf{x}\|_2}
{\|\mathbf{x}'\|_2 + \epsilon}.
\end{equation}
Here, $\|\cdot\|_2$ denotes the Euclidean norm,
and $\epsilon>0$ is a small constant for
numerical stability.

We refer to these two implementations as
\emph{spectral renormalization} and
\emph{spatial renormalization}, respectively.
For simplicity, we collectively denote this step
as $\mathrm{Renorm}$ in Alg.~\ref{alg:swg}.

%% file: sec/4_exp.tex
\section{Experiment}
\label{sec:exp}

\begin{figure*}[t!]
	\centering
	\begin{subfigure}[t]{0.49\textwidth}
		\centering
		\includegraphics[width=\linewidth]{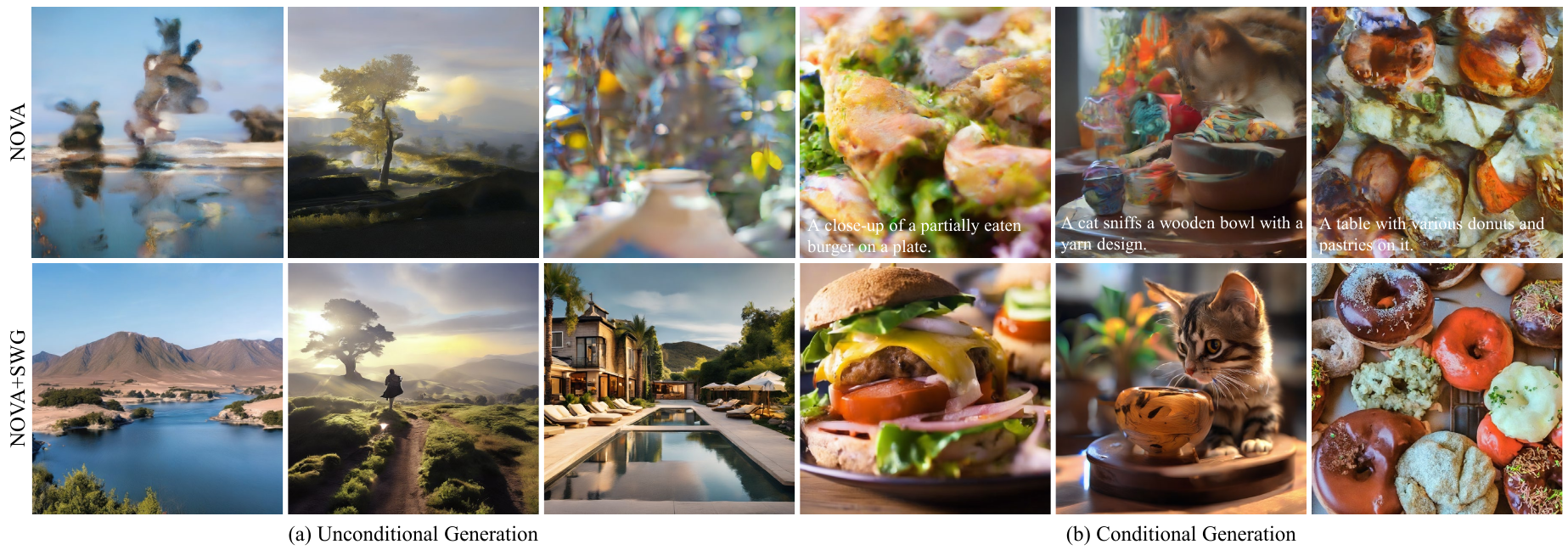}
		\caption{NOVA. The left three columns show null-prompt generations; the right three columns show prompt-conditioned ones. Spatial renormalization is used.}
		\label{fig:main_nova}
	\end{subfigure}
	\hfill
	\begin{subfigure}[t]{0.49\textwidth}
		\centering
		\includegraphics[width=\linewidth]{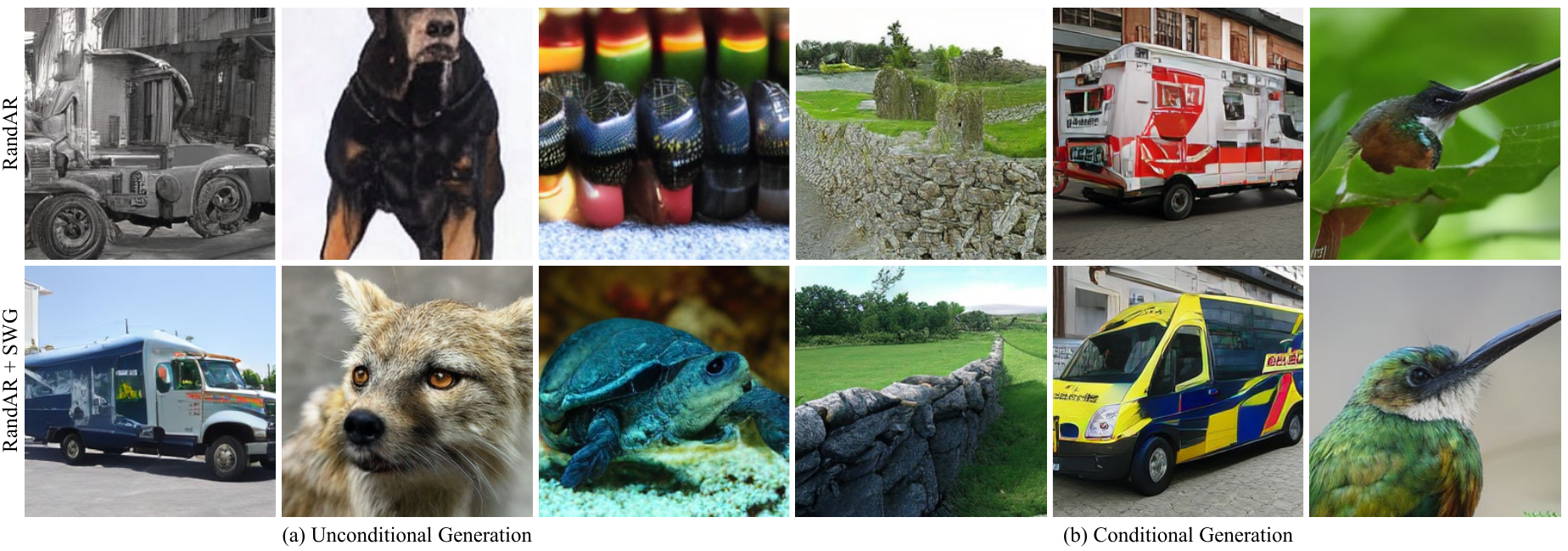}
		\caption{RandAR. The left three columns show unconditional generations; the right three show conditional ones.}
		\label{fig:main_randar}
	\end{subfigure}
	\caption{Visual results of SWG on two autoregressive models.}
	\label{fig:main_visual}
\end{figure*}

\begin{table}[t]
\small
	\centering
    
	\caption{Performance of different visual AR models on the COCO dataset. Both unconditional and conditional generation are considered. CS and AS denote CLIP score and aesthetic score, respectively. 
    }
    \label{tab:main_t2i}
    \begin{subtable}{0.48\linewidth}
    \caption{Unconditional generation}
        \label{tab:main_uncond}
        \begin{tabular}{lccc}
        \toprule[0.1em]
         Method & FID $\downarrow$ & IS $\uparrow$ & AS $\uparrow$ \\
         \midrule[0.1em]
        NOVA & 85.34 & 12.25 & 4.03  \\
        +SSG~\cite{shin2026ssg} & -      & -    & -     \\
        +SWG (Ours) & 50.37 & 19.12 & 5.75 \\ \hline
 Lumina-mGPT & 103.53 & 9.71 & 5.82 \\
        +SSG~\cite{shin2026ssg} & -      & -    & -     \\
        +SWG (Ours) & 101.46 & 9.33 & 6.85 \\ \hline
    Infinity & 142.61 & 5.45 & 5.52 \\
        +SSG~\cite{shin2026ssg} & 157.27 & 4.76 & 5.65   \\
        +SWG (Ours) & 133.45 & 	6.42 & 5.75 \\ 
        \bottomrule[0.1em]  
        \end{tabular}
    % }
    \end{subtable}
    \begin{subtable}{0.48\linewidth}
    \caption{Conditional generation}
        \label{tab:main_cond}
        \begin{tabular}{lcccc}
        \toprule[0.1em]
         Guidance & FID $\downarrow$ & IS $\uparrow$ & CS  $\uparrow$ & AS $\uparrow$ \\
         \midrule[0.1em]
        NOVA & 46.86 & 19.46  & 26.62 & 4.21 \\
        +SSG~\cite{shin2026ssg} & -  & -  & - & -     \\
        +SWG (Ours) & 15.80	& 31.27  & 29.86 & 5.78 \\ \hline
 Lumina-mGPT & 37.33 & 24.50 & 28.07 & 5.98 \\
        +SSG~\cite{shin2026ssg} & -  & -  & - & -  \\
        +SWG (Ours) & 29.91 & 28.43 & 29.28 & 6.70 \\ \hline
    Infinity & 33.92 & 35.43  & 31.21 & 6.23 \\
        +SSG~\cite{shin2026ssg} & 31.49  & 35.84  & 31.22 & 6.24  \\
        +SWG (Ours) & 30.50	& 38.76 & 31.49 & 6.34 \\ 
        \bottomrule[0.1em]
        \end{tabular}
    % } 
    \end{subtable}
\end{table}

\subsection{Experimental Settings}

\noindent
\textbf{Baselines.} 
Our study encompasses random-order, raster-scan, and scale-wise AR paradigms, as shown in Tab.~\ref{tab:bsline}. Since most prior methods rely on the characteristics of diffusion or require access to global spatial structure at every inference step, we include SSG in our comparison of scale-wise AR models. Its original formulation does not directly extend to the other two AR paradigms.

\begin{table}[h]

\centering
\caption{Baseline Model Illustration. Token prediction types: Cont. (continuous) and Disc. (discrete). AR inference paradigms: Rand. (random-order), Rast. (raster-scan), and Scale. (scale-wise).}
\label{tab:bsline}
\begin{tabular}{lccccc}
\toprule[0.1em]
Model & Input & Resolution & \#Params & Token Type & AR Type \\
\midrule[0.1em]
RandAR~\cite{pang2025randar} & Class & 256 & 0.7B & Disc. & Rand. \\
Lumina-mGPT~\cite{liu2024lumina} & Text & 768 & 7B & Disc. & Rast. \\
NOVA~\cite{deng2024autoregressive} & Text & 1024 & 0.3B & Cont. & Rand. \\
Infinity~\cite{han2025infinity} & Text & 1024 & 2B & Disc. & Scale. \\
\bottomrule[0.1em]
\end{tabular}
\end{table}

\noindent
\textbf{Implementation Details.}
By default, we retain 25\% of spectral components, using the same masking strategy across all modified layers. Unless otherwise specified, we apply SWG to layers 4-7 for Lumina-mGPT, RandAR, and Infinity, and to layers 8-11 for NOVA. Experiment-specific settings are provided in the appendix.
The CFG scale (if used) follows their official implementations.

\noindent
\textbf{Datasets and Metrics.} Following common practice~\cite{hong2024smoothed,sadat2024no,rajabi2025token}, we employ COCO dataset~\cite{lin2014microsoft} for text-to-image task evaluation, including both the conditional and unconditional generation settings. 
For class-to-image tasks, we use ImageNet dataset~\cite{deng2009imagenet} for evaluation.
Quantitative results are reported using Fr\'echet inception distance (FID)~\cite{heusel2017gans}, Inception Score (IS)~\cite{salimans2016improved}, CLIP score~\cite{hessel2021clipscore}, Aesthetic score~\cite{schuhmann2022laion}, precision and recall~\cite{kynkaanniemi2019improved}.

\subsection{Main Results}

\begin{figure}[t!]
	\centering
	\includegraphics[width=1.0\linewidth]{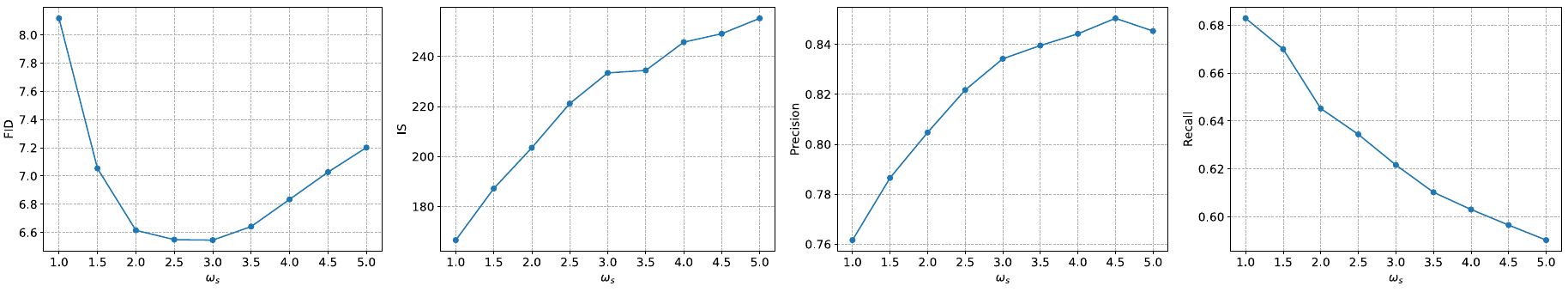}
	\caption{Quantitative evaluation of RandAR under different SWG guidance strength $\omega_s$. From left to right, the curves depict FID, IS, Precision, and Recall with respect to the guidance scale.}
    \label{fig:randar_wg}
\end{figure}

\noindent
\textbf{Qualitative Results.}
We visually illustrate the effectiveness of the proposed SWG in improving the synthesized results.
As shown in Fig.~\ref{fig:main_lumina} and~\ref{fig:main_visual}, these models produce blurred or distorted images without guidance. Moreover, the generated results do not align correctly with the prompts; e.g., `heron' is misinterpreted as `woman' in the $6^{th}$ column in Fig.~\ref{fig:main_lumina}. On the contrary, our proposed SWG significantly improves the quality of synthesized results on both discrete and continuous AR, as well as on unconditional and conditional generation, demonstrating its effectiveness and generalization. 
Note that in unconditional generation scenarios, the reference model itself is the base model, therefore the standard CFG is not applicable.

We also evaluate SWG on class-condition generation. As shown in Fig.~\ref{fig:main_randar}, SWG significantly improves the quality of synthesized results, improving object deformation and blurry outlines.

Additionally, we examine the relationship between the guided and baseline results. Interestingly, the synthesized pairs from NOVA exhibit higher spatial relevance, whereas the pairs from Lumina-mGPT appear to differ. We hypothesize that this phenomenon is attributed to a more compact sampling space of discrete AR models.

\noindent
\textbf{Quantitative Comparisons.}
Our proposed SWG is competent for both unconditional and conditional scenarios on text-to-image generation tasks.
As shown in Tab.~\ref{tab:main_uncond} and Tab.~\ref{tab:main_cond}, SWG improves FID and aesthetic scores across different AR modeling strategies. For the continuous token prediction model NOVA, the gains are particularly pronounced: FID is reduced by 34.97 in unconditional generation and by 31.06 in conditional generation, with simultaneous improvements in terms of text alignment and aesthetics.
For discrete token prediction models, Lumina-mGPT and Infinity, the improvements are more moderate but still consistent. SWG steadily lowers FID and enhances aesthetics in both settings. Specifically, Lumina-mGPT improves CLIP score by 1.21, indicating a stronger ability to follow textual instructions.

Fig.~\ref{fig:randar_wg} shows the effectiveness of our SWG on the class-conditioned generation task. 
As the guidance strength $\omega_s$ increases, the FID score first decreases and then rises, reaching the best value of 6.55 at $\omega_s = 3.0$, indicating an optimal balance between fidelity and diversity. 
The Inception score increases steadily from 166.56 at $\omega_s = 1$ to 255.27 at $\omega_s = 5$, reflecting consistent improvements in perceptual quality. 
Precision also tends to improve overall from 0.76 to around 0.85. Meanwhile, recall gradually decreases from 0.68 to 0.59, suggesting that stronger guidance increases fidelity at the expense of reduced diversity. 
These results indicate that moderate SWG strength yields the most balanced and visually coherent generation quality.

\begin{figure}[t!]
	\centering
	\includegraphics[width=0.95\linewidth]{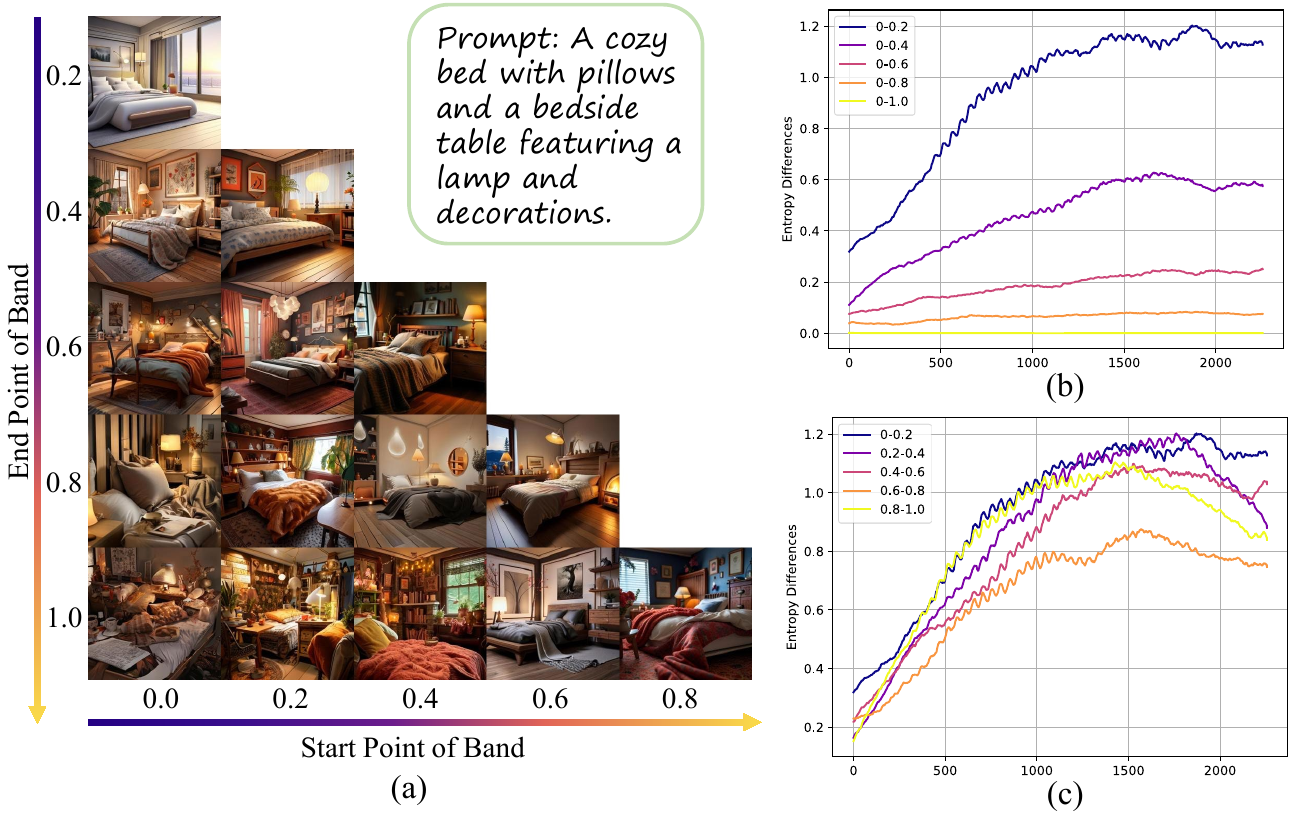}
	\caption{\textbf{Illustration of spectrum analysis.} (a) Visualizations under different spectral band selections. The horizontal axis represents the starting position of the spectral band, while the vertical axis represents the ending position. Bands located between them are retained for the weak model. Apparently, the 0-1.0 interval is equivalent to the baseline without any guidance. \textit{Note that from the bottom left to the top right, the image becomes clearer as the bandwidth decreases.} (b) Entropy difference vs. different bandwidths with the same starting point. (c) Entropy difference vs. different start positions with the same bandwidth. The entropy difference is calculated by subtracting the baseline branch from the perturbation branch. Lumina-mGPT is used for demonstration.}
    \label{fig:abla_sweepband}
\end{figure}

\subsection{Ablation Studies}

\noindent
\textbf{Spectrum Analysis.} 
We consider two key factors regarding spectral bands in weak model construction: bandwidth and location. 
To achieve this, we perform fine-grained sweep ablation. Specifically, we sort the DFT-transformed spectrum according to harmonics, with the entire range denoted as 0-1.0. Then, we set start and end endpoints with a step size of 0.2. Excluding invalid cases where the end point is earlier than or equal to the start point, we obtain a total of 15 images, arranged in a pyramid pattern, as shown in Fig.~\ref{fig:abla_sweepband} (a).

Interestingly, a pattern seems to emerge. The generated results gradually become clearer along the pyramid from the lower left to the upper right (equivalent to from left to right, from bottom to top). However, the generated details seem to decrease (the room becomes neater), suggesting that our \textit{channel-wise perturbation} is fundamentally different from previous \textit{spatial-wise filtering} methods. 

We attribute this phenomenon to the reduction in information.
Due to the varying prediction difficulty across tokens and the cumulative effect of errors, we examine entropy differences to compare the uncertainty of the two branches.
We ablate the bandwidth and position separately, with results shown in Fig.~\ref{fig:abla_sweepband} (b) and (c), respectively. On the one hand, changing the bandwidth has a pronounced effect on the entropy difference. Specifically, as the bandwidth increases from 0.2 to 0.8, the entropy gap between the weak and base branches generally becomes smaller at corresponding inference steps. This trend is consistent with the less pronounced guidance effects observed in the corresponding visual examples. On the other hand, if the bandwidth is kept constant, changing only the location of the retained frequency band has a comparatively smaller effect on the entropy difference, and the curves exhibit broadly similar trends throughout generation.

\begin{figure}[t!]
\centering
\begin{minipage}{0.48\linewidth}
    \centering
    \includegraphics[width=1.0\linewidth]{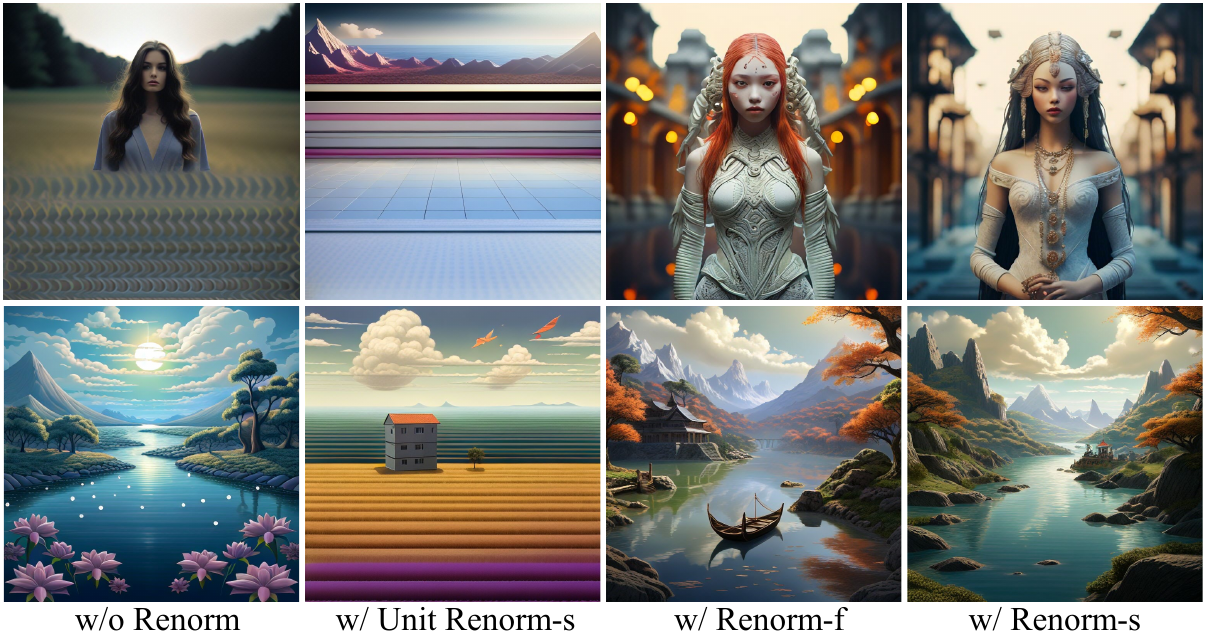}
	\caption{Lumina-mGPT performs unconditional generation under different renormalization strategies. Renorm-f and Renorm-s indicate the renormalization in spectral and spatial domain, as introduced in Eq.~\ref{eq:spec_renorm} and Eq.~\ref{eq:spat_renorm}, respectively. Unit Renorm-s means use spatial normalization to unit length.}
    \label{fig:ab_renorm}
\end{minipage}
\hfill
\begin{minipage}{0.48\linewidth}
    \centering
    \includegraphics[width=1.0\linewidth]{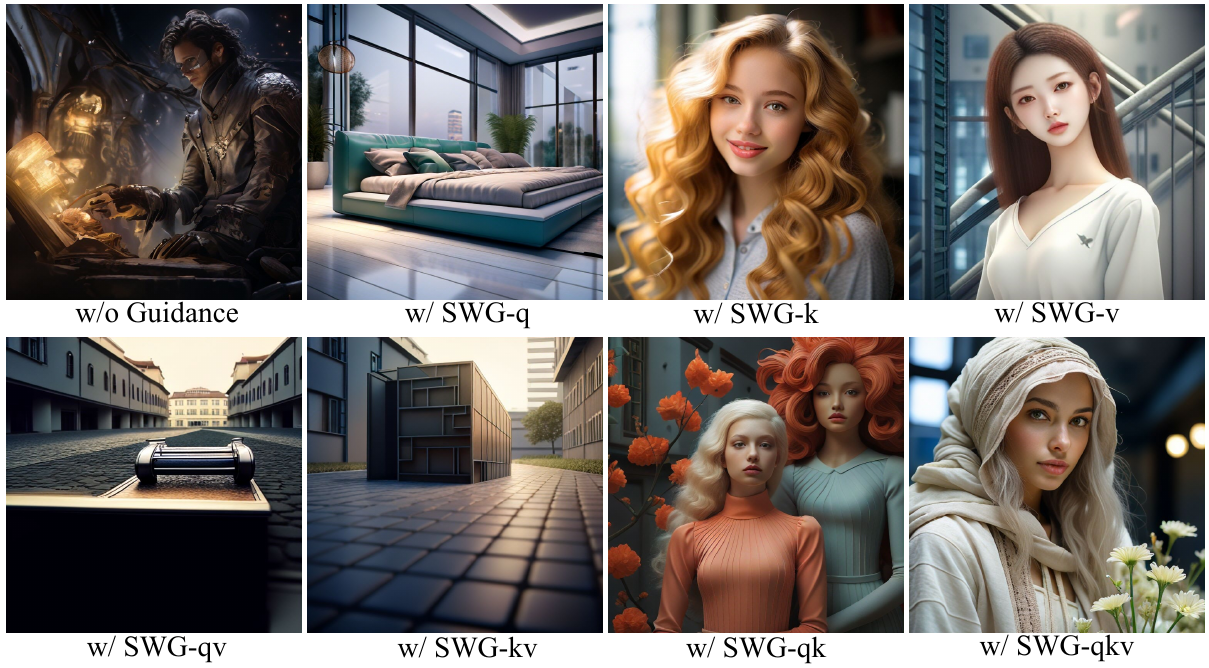}
	\caption{Lumina-mGPT performs unconditional generation employing various SWG strategies within the attention module. The term w/o guidance denotes the absence of any guidance mechanism. Configurations associated with SWG specify whether the query (Q), key (K), and value (V) representations are processed independently or jointly.}
    \label{fig:ab_lumina_qkv}
\end{minipage}
\end{figure}

\noindent
\textbf{Ablation of Renormalization.}
As shown in Fig.~\ref{fig:ab_renorm}, different signal renormalization strategies yield varying synthesis results. 
Applying renormalization, whether in the spatial or spectral domain, generally enhances the quality of the generated outputs, with subtle differences in layout and texture. 
In contrast, omitting this step results in repetitive patterns and can even cause the model to fail as the generation nears completion. 
This issue is primarily due to error accumulation in the AR models, particularly in the raster-scan paradigm. Furthermore, increasing the gap between the original and processed signals exacerbates this problem, resulting in visible artifacts, such as repetitive lines, as shown in the second column of the figure.

\begin{table}[t]
\centering
\caption{Comparison of unconditional generation on COCO using NOVA, evaluated on 1K generated images. AS denotes aesthetic score. \textbf{Bold} indicates the best entries. Spatial and Spectrum indicate SWG with spatial and spectral renormalization, respectively.
}
\label{tab:ab_variant}
\begin{tabular}{lccccc}
\toprule[0.1em]
 & Baseline & Average & Pruning & Spatial & Spectrum \\
\midrule[0.1em]
FID $\downarrow$ & 139.75 & 112.79 & 110.87 & 105.98 & \textbf{104.51} \\
IS $\uparrow$    & 9.55   & 11.87  & 13.25  & 13.04  & \textbf{13.44} \\
AS $\uparrow$    & 4.01   & 5.56   & 5.02   & 5.70   & \textbf{5.72} \\
\bottomrule[0.1em]
\end{tabular}
\end{table}

\noindent
\textbf{Ablation of Weak Branch Constructions.}
In addition to SWG, we design some other weak models and compare them in Tab.~\ref{tab:ab_variant}. Naively averaging the components or pruning specific layers has a limited effect on guiding. Specifically, these operations yield suboptimal results, with FID scores of 112.79 and 110.87, respectively. In contrast, SWG consistently outperforms on these metrics.

SWG demonstrates robustness to perturbations in the positional components. 
As illustrated in Fig.~\ref{fig:ab_lumina_qkv}, the model fails for unconditional generation results without guidance, and the synthesized outputs appear blurry and distorted.
In contrast, integrating SWG substantially enhances the visual quality of the generated results. SWG remains effective regardless of whether it is applied to the query, key, or value representations.

Interestingly, different perturbation locations can be combined flexibly. 
While these settings lead to changes in layout or contextual emphasis, the resulting output remains coherent and visually plausible. 
This consistency demonstrates the benefits of our spectrum-weakening design. The operation preserves the structural stability of intermediate representations, allowing the model to process them smoothly. 
It attenuates certain spectral components in the intermediate representations, yielding a weaker model variant that aligns with our view that the guidance effect does not rely on a particular structural configuration.

\begin{figure}[t]
	\centering
	\includegraphics[width=0.9\linewidth]{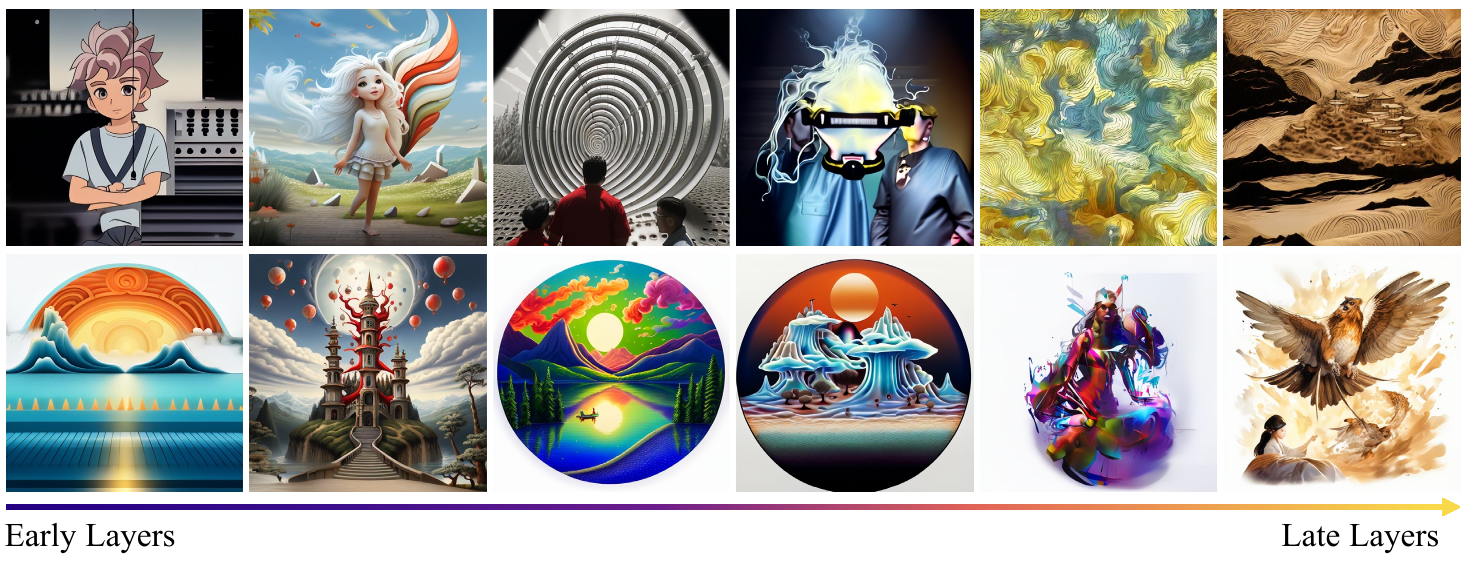}
	\caption{Applying SWG on different layers. Lumina-mGPT with a null prompt is used for demonstration. As the index of layers increases, the effectiveness of guidance decreases, and the synthesized results become blurry and distorted.}
    \label{fig:abla_layer}
\end{figure}

\noindent
\textbf{Layer Selection.}
Fig.~\ref{fig:abla_layer} illustrates the impact of SWG application on the generated results at different layers. Intuitively, the higher the index number, the fewer layers it can affect, and the lower the guiding effect. Notably, applying SWG at the very beginning of the model seems to cause excessive differences between the weak model and the base model, resulting in a lack of detail in the final generated result, consistent with observations in Fig.~\ref{fig:abla_sweepband}.

%% file: sec/5_conclusion.tex
\section{Conclusion}
\label{sec:conclusion}
This paper introduces SWG, a training-free visual guidance framework that builds a weak model by selectively retaining a subset of spectral components in intermediate representations, with a renormalization step to maintain feature scales and support stable generation.
Demonstrating effective performance across various unconditional and conditional settings and AR modeling tasks, SWG extends the scope of CFG, offering a broader and more flexible paradigm for generation control.

\noindent
\textbf{Acknowledgment.} This work was supported by Shenzhen Science and Technology Innovation Project (Grant No.2025110105) and NSFC (Grant No.42271426).

%% file: sec/6_suppl.tex
\newpage
\appendix
\noindent
\textbf{Overview.} In the supplementary materials, we present more results and details:
\begin{itemize}
    \item \textbf{\ref{sec:appendix_details}.} More details on the method. 
    \item 
    \textbf{\ref{sec:appendix_studies}.}
    More empirical studies and visual results.
    \item 
    \textbf{\ref{sec:appendix_limit}.} Discussion of the broader impacts, safeguards and limitations.
    \item 
    \textbf{\ref{sec:appendix_compute}.} 
    Compute Reporting.
\end{itemize}

\section{Method Details}
\label{sec:appendix_details}

\noindent \textbf{Implementation Details.}
We apply spectrum selection to the attention module. By default, we retain the components located in the 0.25-0.5 spectral band.
The guidance strength is set to 2 for Lumina-mGPT, 3 for NOVA, and 4 for Infinity, respectively.
We use Qwen2.5-VL to generate captions for COCO. Unless otherwise specified, we evaluate on 30K generated images for COCO and compute FID on 10K generated images for ImageNet.

\noindent \textbf{Reason for Renormalization.}
Recall that SWG's core operation is $\mathbf{y}=\mathbf{W}^{-1}\mathbf{M}\mathbf{W}\mathbf{x}$, with $\mathbf{W}$ and $\mathbf{M}$ defined in Eqs.~\ref{eq:dft_matrix_single} and~\ref{eq:sel_matrix}, respectively. According to Parseval's theorem, we obtain $\|\mathbf{y}\|_2=\|\mathbf{M}\mathbf{W}\mathbf{x}\|_2\leq\|\mathbf{W}\mathbf{x}\|_2=\|\mathbf{x}\|_2$, with equality when the discarded spectral components have zero energy.

\section{Empirical Results}
\label{sec:appendix_studies}

\subsection{More Results on Class-condition Models}

In this subsection, we provide more results on class-condition models, including random-order and scale-wise prediction.

\begin{figure}[ht]
	\centering
	\includegraphics[width=0.8\linewidth]{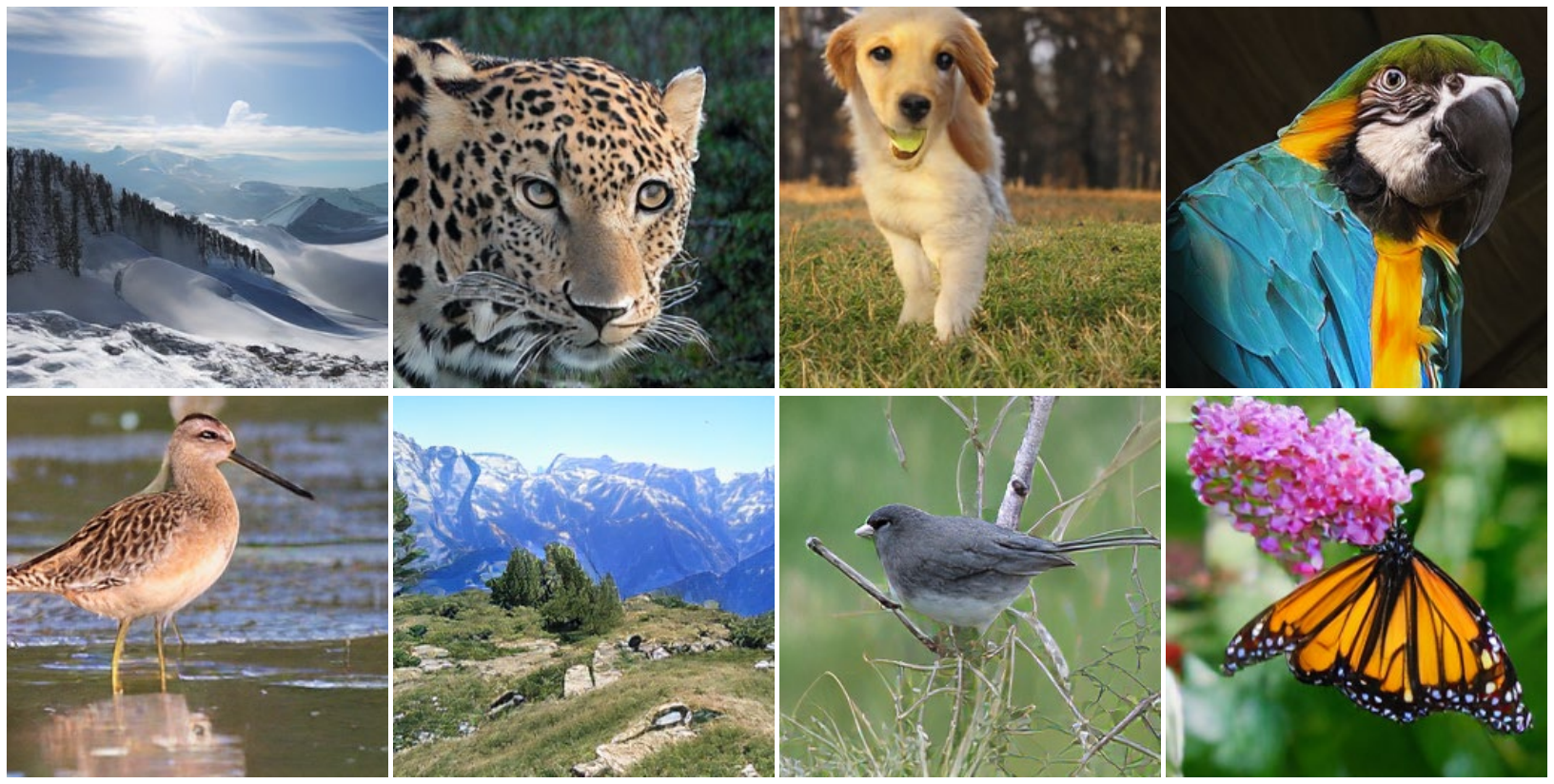}
	\caption{More visual results of RandAR on ImageNet dataset.}
    \label{fig:sup_randar}
\end{figure}

For RandAR evaluation with 50K images, we perturb layers 8 and 9, using CFG and SWG scales of 4 and 0.2, respectively.
As shown in Tab.~\ref{tab:sup_randar_50k}, SWG is compatible with CFG and can further enhance the performance. Specifically, our SWG further improves the FID to 2.23. More visual results are in Fig.~\ref{fig:sup_randar}.

For VAR, we apply SWG on the layer 0-3 on VAR-d16, and set the guidance scale 2.0 for unconditional generation and 5.5 for conditional generation, respectively. As shown in Tab.~\ref{tab:sup_var}, our proposed SWG consistently improves generation quality on scale-wise AR models.

\begin{table}[ht]
\centering
\caption{Performance of RandAR on class-conditional ImageNet $256\times256$ benchmark. 50K images are generated for evaluation, with a sampling step of 88.}
\label{tab:sup_randar_50k}
\begin{tabular}{lcccc}
\toprule[0.1em]
 & FID $\downarrow$ & IS $\uparrow$ & Precision $\uparrow$ & Recall $\uparrow$ \\
 \midrule[0.1em]
 RandAR w/ CFG & 2.25 & 317.77 & 0.80 & 0.60 \\
+SWG & 2.23 & 303.96 & 0.81 & 0.59 \\
\bottomrule[0.1em]
\end{tabular}
\end{table}

\begin{table}[ht]
    \caption{Performance of VAR-d16 on class-conditional ImageNet $256\times256$ benchmark. 10K images are generated for evaluation on ImageNet.}
    \centering
    \label{tab:sup_var}
    \begin{tabular}{lccccc}
        \toprule[0.1em]
         & Guide & FID $\downarrow$ & IS $\uparrow$ & Prec. $\uparrow$ & Rec. $\uparrow$ \\
         \midrule[0.1em]
        \multirow{2}{*}{Uncond} & - & 54.43 & 19.19 & 0.53 & 0.47 \\
         & SWG  & 41.96 & 23.34 & 0.58 & 0.45 \\ \hline
        \multirow{2}{*}{Cond} & - & 14.49 & 104.15 & 0.71 & 0.66 \\
         & SWG & 8.74 & 188.24 & 0.82 & 0.58 \\
        \bottomrule[0.1em]
        \end{tabular}
\end{table}

\subsection{More Results on Text-condition Models}

\begin{figure}[t]
	\centering
	\includegraphics[width=1.0\linewidth]{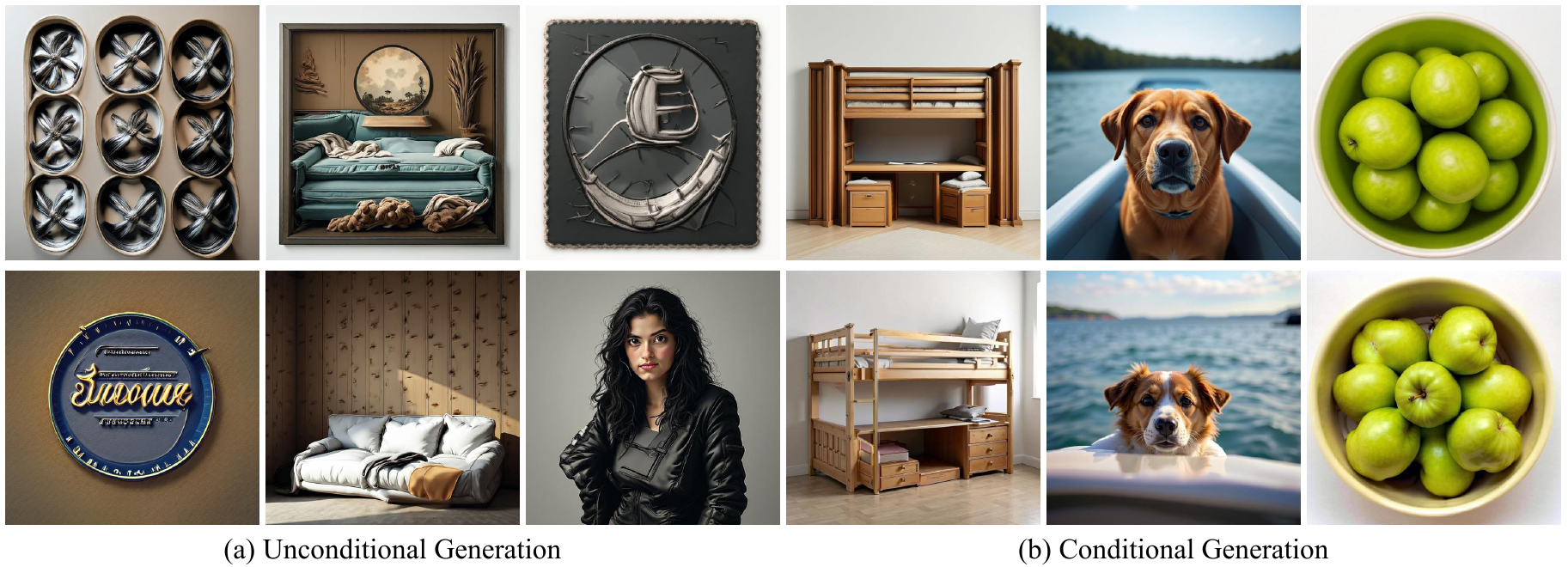}
	\caption{More visual results of Infinity on COCO datasets. (a) Unconditional generation with null prompts. (b) Conditional generation with specific prompts. The prompts from left to right: ``\textit{A wooden bunk bed with a desk and drawers underneath}'', ``\textit{Dog peeking out of a pontoon boat}'', ``\textit{Bright green apples in a bowl}''.}
    \label{fig:sup_infi}
\end{figure}

In this subsection, we provide more results on text-to-image models.
Fig.~\ref{fig:sup_infi} visualizes the effectiveness of our proposed SWG on Infinity. The synthesized results exhibit better detail and realism. Moreover, SWG can effectively improve the DPG-Bench score, as shown in Tab.~\ref{tab:infi_dpg}.

\subsection{Inference Time Analysis}

We analyzed the inference time. Due to DFT, our proposed SWG is slightly slower than CFG. However, since this operation is applied only to specific layers, the additional time cost is limited, as shown in Tab.~\ref{tab:sup_speed}.

\begin{table}[t]
\centering
\caption{Inference time comparison between SWG and CFG. RandAR 0.7B is used for evaluation with a batch size of 64.}
\label{tab:sup_speed}
\begin{tabular}{lcc}
\toprule[0.1em]
 & CFG & SWG \\
 \midrule[0.1em]
images / sec. & 12.59 & 11.62 \\
\bottomrule[0.1em]
\end{tabular}
\end{table}

\begin{table}[t]
\centering
\caption{Evaluation of SWG on Infinity using DPG-Bench. We report the overall score.}
\label{tab:infi_dpg}
\begin{tabular}{lcc}
\toprule[0.1em]
 & - & SWG \\
 \midrule[0.1em]
Overall & 73.43 & 77.03 \\
\bottomrule[0.1em]
\end{tabular}
\end{table}

\begin{figure}[ht]
\centering
\begin{minipage}{0.48\linewidth}
    \centering
    \includegraphics[width=0.7\linewidth]{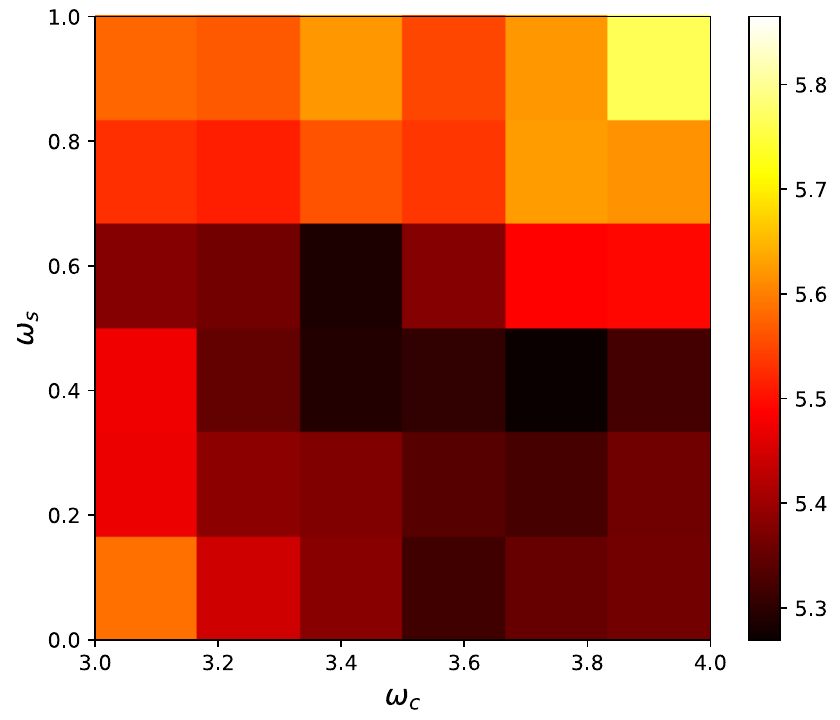}
	\caption{Compatibility between SWG ($\omega_s$) and CFG ($\omega_c$) on the ImageNet dataset. The heatmap shows FID scores, with darker colors indicating better performance.}
    \label{fig:sweep_c2i}
\end{minipage}
\hfill
\begin{minipage}{0.48\linewidth}
\small
\captionsetup{type=table}
    \centering
    \caption{Compatibility of SWG and CFG on COCO using NOVA, evaluated on 1K generated images. CS and AS indicate CLIP score and aesthetic score, respectively. \textbf{Bold} indicates the best entries.}
\label{tab:ab_cfg}
\begin{tabular}{cccccc}
\toprule[0.1em]
 Method & FID $\downarrow$ & IS $\uparrow$ & CS $\uparrow$ & AS $\uparrow$ \\
\midrule[0.1em]
 NOVA w/CFG & {67.59} & {25.67} & \textbf{32.86} & {5.80} \\
 +SWG & \textbf{67.02} & \textbf{26.30} & {32.51} & \textbf{5.96} \\
\bottomrule[0.1em]
\end{tabular}
    
\end{minipage}
\end{figure}

\begin{table}[ht]
\centering
\caption{Compatibility of SWG and CFG on GenEval.}
\label{tab:sup_infi_geneval}
\begin{tabular}{cccccccc}
\toprule
Method & Single Obj & Two Obj & Counting & Colors & Pos & Color Attr & Overall \\
\midrule
Infinity w/CFG & 1.00 & 0.76 & 0.62 & 0.84 & 0.25 & 0.58 & 0.67 \\
+SWG           & 1.00 & 0.76 & 0.60 & 0.86 & 0.27 & 0.62 & 0.68 \\
\bottomrule
\end{tabular}
\end{table}

\begin{figure}[ht]
\centering
\begin{minipage}{0.4\linewidth}
    \centering
    \includegraphics[width=1.0\linewidth]{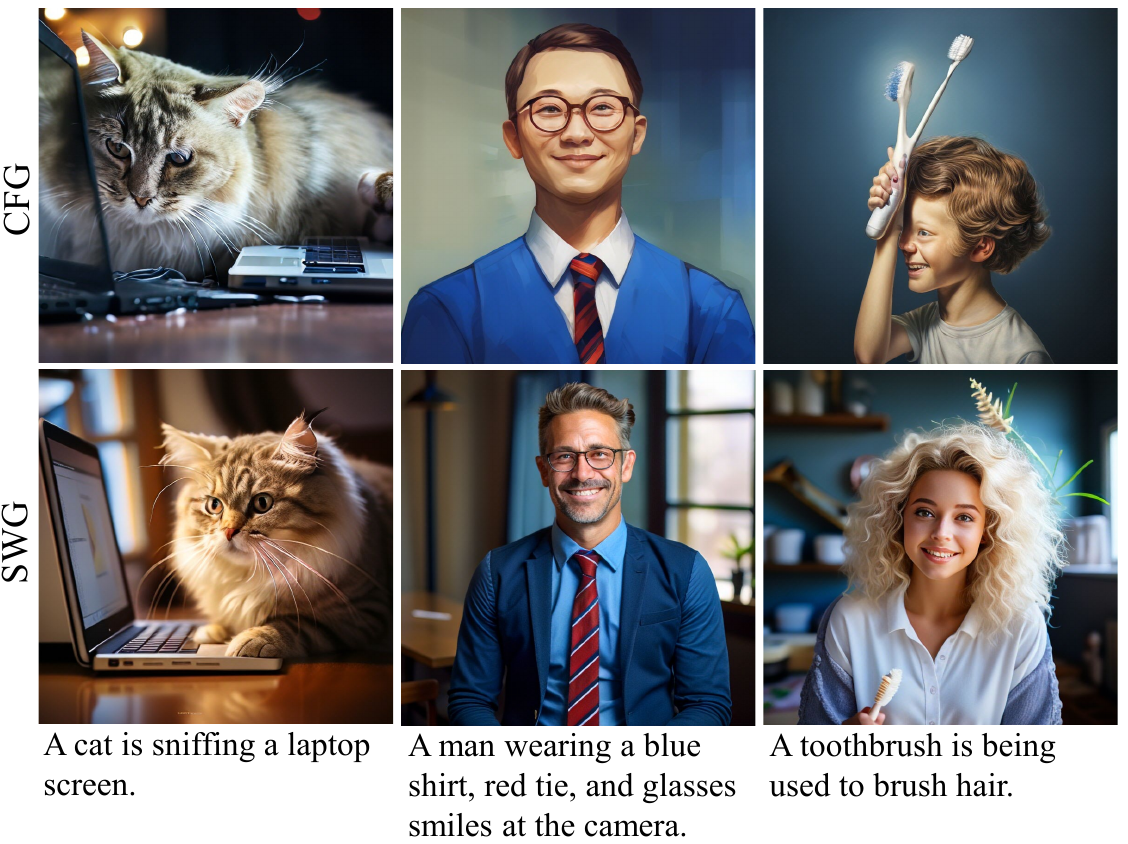}
	\caption{Visualization of different guidance strategies on Lumina-mGPT. Compared to CFG, SWG is superior in overall image clarity and aesthetic quality.}
    \label{fig:lumina_cfg}
\end{minipage}
\hfill
\begin{minipage}{0.55\linewidth}
\small
\captionsetup{type=table}
    \centering
    \caption{Comparison with CFG on COCO using Infinity, evaluated on 1K generated images. CS and AS indicate CLIP score and aesthetic score, respectively. Uncond. Train indicates whether guidance requires unconditional training support.}
\label{tab:ab_cfg_infi}
\begin{tabular}{cccccc}
\toprule[0.1em]
Uncond. Train & Guide & FID $\downarrow$ & IS $\uparrow$ & CS $\uparrow$ & AS $\uparrow$ \\
\midrule[0.1em]
- & - & 79.28 & 22.84 & 31.45 & 6.24 \\
\checkmark & CFG & 77.17 & 25.22 & 32.29 & 6.25 \\
\ding{55} & SWG & 74.69 & 24.06 & 31.73 & 6.39 \\
\bottomrule[0.1em]
\end{tabular}
    
\end{minipage}
\end{figure}

\noindent
\textbf{Compatibility with CFG.}
We demonstrate both quantitatively and visually that SWG is compatible with CFG.
As shown in Tab.~\ref{tab:ab_cfg} and Tab.~\ref{tab:sup_infi_geneval}, SWG is compatible with CFG for enhanced aesthetic score and overall GenEval performance.
Fig.~\ref{fig:sweep_c2i} further supports this finding, where the optimal configuration attains an FID of 5.26 at $\omega_c=3.8, \omega_s=0.4$, surpassing either component used alone.

Interestingly, while both SWG and CFG provide effective guidance, they emphasize different aspects.
In Tab.~\ref{tab:ab_cfg_infi}, compared with the baseline method, both CFG and our SWG can consistently improve the generation quality. However, CFG yields the highest CLIP score, reflecting more substantial prompt alignment, whereas SWG attains higher aesthetic scores, placing greater emphasis on highlighting image sharpness and overall quality.
Note that our proposed SWG is training-free, while CFG relies on unconditional predictions learned during training as the reference for guidance.
Qualitative examples in Fig.~\ref{fig:lumina_cfg} also show that the images synthesized by SWG have richer details and are more realistic.

\subsection{Extension to Diffusion Models}

\begin{figure*}[ht]
	\centering
	\includegraphics[width=0.8\linewidth]{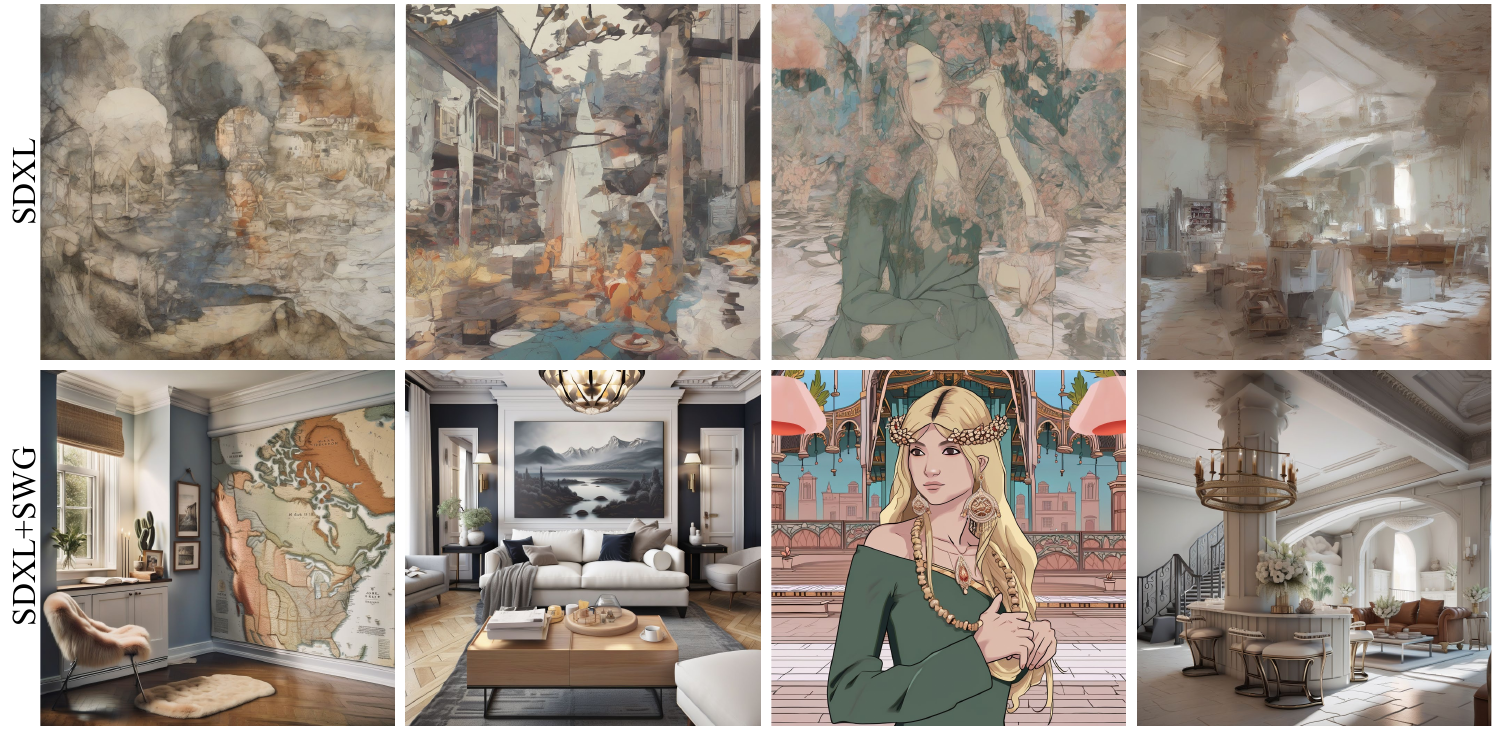}
	\caption{Visual results of applying SWG to SDXL with an empty prompt.}
    \label{fig:sup_sdxl}
\end{figure*}

SWG also extends to diffusion models. We evaluate it on SDXL by applying spectrum selection to query representations in the middle layers and retaining the last 10\% of spectral components. Fig.~\ref{fig:sup_sdxl} shows improved visual quality with an empty prompt, while Tab.~\ref{tab:sup_sdxl} reports competitive performance against Smoothed Energy Guidance (SEG) in both unconditional and conditional generation.

\begin{table}[ht]
    \centering
    \caption{Comparison with diffusion baselines on COCO dataset. SDXL is used for demonstration.}
    \label{tab:sup_sdxl}
    \begin{tabular}{lccccc}
        \toprule[0.1em]
         & Guide & FID $\downarrow$ & IS $\uparrow$ & CS  $\uparrow$ & AS $\uparrow$ \\
         \midrule[0.1em]
        \multirow{3}{*}{Uncond} & - & 185.76 & 7.23 & - & 4.82 \\
         & SEG  & 138.14 & 9.90 & - & 6.09 \\ 
         & SWG  & 137.75 & 10.67 & - & 6.14 \\ \hline
        \multirow{3}{*}{Cond} & - & 108.03 & 14.58 & 27.60 & 5.20 \\
         & SEG  & 75.02 & 20.06 & 30.12 & 6.14 \\ 
         & SWG & 68.94 & 23.89 & 32.72 & 6.13 \\
        \bottomrule[0.1em]
        \end{tabular}
\end{table}

\begin{figure}[ht]
	\centering
	\includegraphics[width=0.8\linewidth]{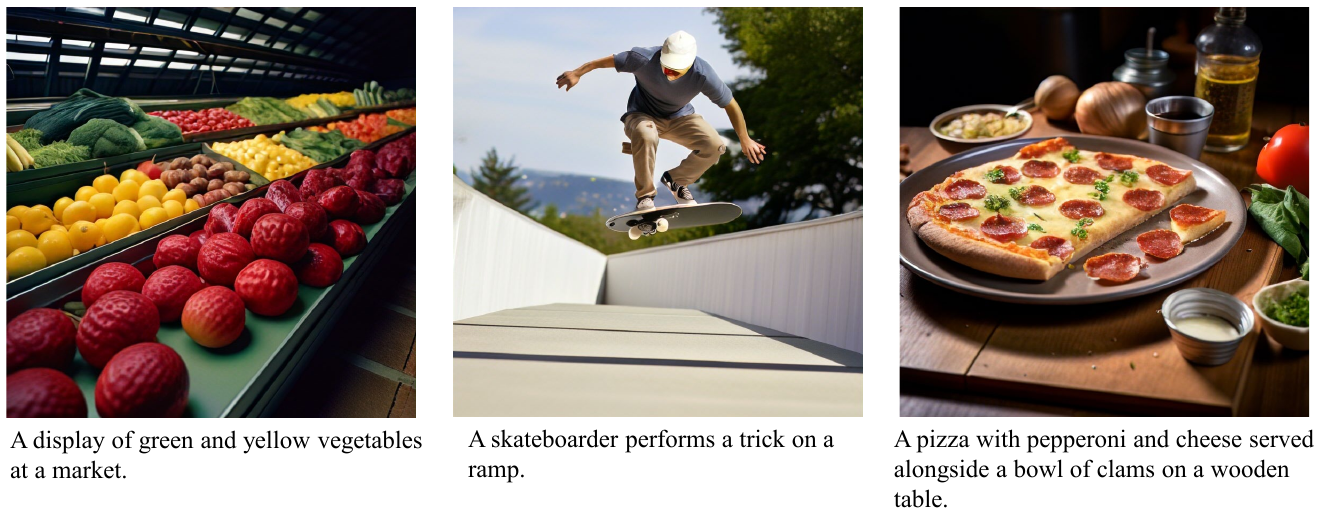}
	\caption{Failure case analysis of Lumina-mGPT. The prompts corresponding to each generated result are provided below.}
    \label{fig:sup_failcase}
\end{figure}

\subsection{Failure Cases}

Our proposed method occasionally produces artifacts or fails to fully adhere to the instructions, as illustrated in Fig.~\ref{fig:sup_failcase}.

\section{Discussion}
\label{sec:appendix_limit}

\subsection{Broader Impacts}

This paper proposes a training-free and condition-free guidance approach for visual AR generation that improves image quality without retraining the underlying models.
The method may benefit visual content creation, but enhanced image realism may also facilitate the creation of misleading content.
As SWG operates on pretrained models, its outputs may inherit biases and limitations from the underlying training data.

\subsection{Safeguards}

Our method is training-free, which means it does not introduce new knowledge into the pre-trained models. Since training data heavily influences the generated content, ensuring its fairness and cleanliness is crucial to preventing harmful content.

\subsection{Limitations}

Despite the promising results of our method, it still has limitations. SWG sometimes fails to capture fine details or has artifacts in the generated results, suggesting room for improvement.

\section{Compute Reporting}
\label{sec:appendix_compute}

All experiments are conducted on a single-node cluster equipped with Intel(R) Xeon(R) Gold 6248 CPUs @ 2.50GHz, 8 NVIDIA A100 GPUs, and 1.5 TB of system memory. Our method requires no training.